\documentclass{article}

\usepackage{comment}
\usepackage{microtype}
\usepackage{graphicx}
\usepackage{subfigure}
\usepackage{booktabs} 
\usepackage{multirow}
\usepackage{adjustbox}
\usepackage{hyperref}

\usepackage[accepted]{icml2025}

\usepackage{amsmath}
\usepackage{amssymb}
\usepackage{mathtools}
\usepackage{amsthm}

\usepackage{multirow}

\usepackage[capitalize,noabbrev]{cleveref}

\theoremstyle{plain}

\theoremstyle{definition}

\theoremstyle{remark}

\usepackage{enumitem}

\usepackage[textsize=tiny]{todonotes}

\icmltitlerunning{Huff-LLM: End-to-end lossless compression for LLMs}

\begin{document}

\twocolumn[
\icmltitle{Huff-LLM: End-to-End Lossless Compression for Efficient LLM Inference}



\icmlsetsymbol{equal}{*}

\begin{icmlauthorlist}
\icmlauthor{Patrick Yubeaton}{equal,nyu}
\icmlauthor{Tareq Mahmoud}{equal,und}
\icmlauthor{Shehab Naga}{equal,und}
\icmlauthor{Pooria Taheri}{equal,und}
\icmlauthor{Tianhua Xia}{equal,nyu}
\icmlauthor{Arun George}{und}
\icmlauthor{Yasmein Khalil}{und}
\icmlauthor{Sai Qian Zhang}{nyu}
\icmlauthor{Siddharth Joshi}{und}
\icmlauthor{Chinmay Hegde}{nyu}
\icmlauthor{Siddharth Garg}{nyu}
\end{icmlauthorlist}
\icmlaffiliation{nyu}{Department of Electrical and Computer Engineering, New York University, NY, USA}

\icmlaffiliation{und}{Department of Computer Science and Engineering, University of Notre Dame, IN, USA}


\icmlcorrespondingauthor{Patrick Yubeaton}{wpy2004@nyu.edu}

\icmlkeywords{Machine Learning, ICML}

\vskip 0.3in
]



\printAffiliationsAndNotice{\icmlEqualContribution} 

\begin{abstract}
As they become more capable, large language models (LLMs) have continued to rapidly increase in size. This has exacerbated the difficulty in running state of the art LLMs on small, edge devices. Standard techniques advocate solving this problem through lossy compression techniques such as quantization or pruning. However, such compression techniques are lossy, and have been shown to change model behavior in unpredictable manners. We propose Huff-LLM, an \emph{end-to-end, lossless} model compression method that lets users store LLM weights in compressed format \emph{everywhere}---cloud, disk, main memory, and even in on-chip memory/buffers. This allows us to not only load larger models in main memory, but also reduces bandwidth required to load weights on chip, and makes 
more efficient use of on-chip weight buffers. In addition to the memory savings achieved via compression, we also show latency and energy efficiency improvements when performing inference with the compressed model.
\end{abstract}

\section{Introduction}


State-of-art Large language models (LLMs) are \emph{massive}---even a mid-size model like Llama3-70B~\cite{dubey2024llama} takes up 150GB of memory, which is out of reach except for the highest-end hardware. 
Model size not only limits deployment on edge devices that tend to have small memory capacity, but also 
increases the memory bandwidth required for fast inference. 
Massive model sizes have
motivated a large body of work on model compression targeted specifically for LLMs. 
These methods primarily fall into two buckets: quantization and pruning. Quantization methods seek to decrease the precision of model parameters, thus requiring fewer bits per parameter, while pruning methods seek to decrease the number of parameters. 
Recent works such as LLM.Int8()~\cite{dettmers2022gpt3}, GPTQ~\cite{frantar2022gptq}, and AWQ~\cite{lin2024awq} have achieved 2$\times$-4$\times$ compression with respect to LLM weights, thereby potentially speeding up inference.  

However, quantization and pruning are both lossy compression methods, leading the compressed 
models to behave differently from the original model and flipping incorrect to correct answers (and vice versa) on multiple-choice benchmarks even if average accuracy is maintained~\cite{dutta2024accuracy}.  
Safety, trustworthiness, multilingual capabilities and demographic biases might also be impacted, as shown by  \cite{xu2024beyond,hong2024decoding,marchisio2024does}. 
%
These results demonstrate that, notwithstanding the impressive results obtained from lossy model compression, we have yet to fully understand its impact on LLM behaviour.  
This raises the question: 
\emph{can we compress LLMs without altering their behaviour in any way}?

Lossless compression methods (such as Huffman coding and arithmetic coding) offer a solution. 
Just as how a Huffman-compressed image can be reconstructed 
{exactly} in its original form;  
a losslessly compressed LLM model would behave \emph{identically} to the original model after decompression.
However, despite widespread use in other domains, 
lossless compression has found surprisingly little application in LLM compression. 
One main reason is that 
lossless compression and decompression can be  \emph{computationally expensive} and is not natively supported on commodity hardware like CPUs and GPUs. Custom hardware accelerators (such as TPUs and NPUs) also do not implement lossless compression due to hardware implementation overheads.

Prior work has proposed 
lossless compression to reduce download costs of LLM weights from the cloud~\cite{hershcovitch2024zipnn}, but the model is decompressed and loaded into memory
in its original, 
uncompressed format.
~\cite{hao2024neuzip} go a step further: models are loaded into memory in compressed form, but decompressed layer by layer during inference.
Thus, larger models can be loaded into a smaller main memory, but at the cost of \emph{increased} inference latency since weight matrices must first be decompressed before inference.  
As a result, prior methods do not  realize the \emph{full} benefits of model compression, including reduced download costs and memory footprint, 
but also
faster and more energy-efficient LLM inference.


\paragraph{Our contributions.} 
We propose \textsc{Huff-LLM}, a new \emph{end-to-end} model compression 
method and custom hardware implementation that stores LLM weights in compressed format \emph{everywhere}---cloud, disk, main memory, \emph{and} in on-chip memory/buffers. Weights are only decompressed when needed, \emph{i.e.}, to multiply with inputs/activations, where they are decompressed to their original FP16/BF16 formats. Via careful hardware-software co-design, we ensure that \textsc{Huff-LLM} is both lightweight, adding less than 6\% area overhead, and easily integrated into custom hardware architectures like systolic arrays and vector-accelerator~\cite{nvdla, simba, nvidia_xformer} architectures commonly
used in today's TPU/NPU chips. 
Using simulations and an FPGA prototype, we show that \textsc{Huff-LLM}
reduces model size by up to 32\%, improves
inference latency by up to 31\%, \emph{and} cuts energy cost by up to 26\%. 
Our main contributions 
are as follows:
\begin{itemize}[nosep,leftmargin=*]
\item We introduce \textsc{Huff-LLM}, an end-to-end model compression technique which is capable of maintaining LLM weights in compressed format throughout the system when using custom hardware.
\begin{itemize}
    \item Rather than applying Huffman compression to the whole parameter, \textsc{Huff-LLM} compresses subsets of LLM weight parameters. This minimizes the overheads of Huffman decompression, rendering it practical.
    \item We develop a Huffman decoder that can, with minimal overhead, be integrated into standard accelerator architectures like Systolic Arrays and Vector Processors.
\end{itemize}
\item Our evaluations across multiple LLM architectures demonstrate that \textsc{Huff-LLM} achieves a 15--32\% reduction in both required on-chip memory capacity and memory bandwidth requirements.
\item We evaluate \textsc{Huff-LLM}'s compression performance using accelerator design tools, simulations, and standard performance estimators across multiple accelerator architectures and popular open-weight LLM families.
\begin{itemize}
    \item We observe consistent savings across architectures, with up to 31\% improvement to latency and up to 26\% reduction in energy.
\end{itemize}
\end{itemize}


\section{Background and Related Work}

\subsection{LLM Model Compression}


\paragraph{LLM Weight Formats.} LLM weights are broadly stored in either floating point or integer formats during inference.
The most common floating point formats are 32-bit floating point (FP32), 16-bit floating point (FP16), and 16-bit brain float (BF16). 
In FP16, for instance, the most significant bit (MSB) is the {sign} (S) of the number, the next five bits are the {exponent} (E), and the last ten bits are the {mantissa} (M). 
Any FP16 weight is then represented as: 
\begin{equation}
    \text{Value} = (-1)^S \times 2^{E-B} \times (1.M)
\end{equation}
where $B$ is a fixed bias term, commonly set to $B=15$. In this format, mantissa bits only encode 
the fractional value after the decimal point ($1.M$).
FP32 is similar, but allocates 8 bits to the exponent and 23 bits to the mantissa. 
BF16, introduced as a compromise between FP32 and FP16, has an 8-bit exponent, 
a 7-bit mantissa, and a sign bit. 
BF16 has a larger dynamic range compared to FP16, but lower precision within this range compared
to FP16.

To reduce model size, 8-bit integer (INT8) and 4-bit integer (INT4) formats were introduced.
These represent weights as 
signed 2's-complement integers. Decimals are represented using 
a scaling factor typically associated with an entire tensor or channels within it.
INT8 and INT4 representations are typically obtained via quantization methods applied to models 
stored in FP16 or BF16 formats. These methods are discussed next.

\paragraph{LLM Weight Quantization} 
\textcolor{black}{As LLM sizes have grown, their associated workloads often become expensive in real-world applications. Techniques to reduce the memory footprint and computation precision for these models has been driven by the need to serve these models at scale, or run larger models locally on compute-limited resources. Quantization-aware training (QAT), has demonstrated the highest accuracy for most models; in QAT, precision reduction through quantization is included within the training loop, often requiring that the training pipeline and training data be accessed during the quantization process. Usually, QAT is incorporated within the fine-tuning phase~\cite{yao2022zeroquant}, resulting in significant training overhead cost; these costs are particularly exacerbated for large-scale models. Post-training Quantization (PTQ), in contrast, quantizes existing pre-trained models, avoiding the need to retrain the model. The PTQ approach avoids many privacy hurdles associated with access to pre-training, enabling third parties to modify open-weight models and serve them efficiently on various hardware platforms~\cite{lie2023cerebras,frantar2022gptq, ashkboos2024quarot}.}

However, while post-training quantization can reduce inference costs, reducing model precision is often associated with unintended consequences~\cite{sambanova2023reducedprecision, cerebras2024llama3}. Research such as~\cite{zhang2024does} has shown that post-hoc quantization can have adverse effects on model alignment, and can be used to mitigate ``unlearning'' procedures that are applied to LLMs as copyright or safety filters. Quantization applied to multi-lingual LLMs have disparate effects on low-resource languages, particularly those that use non-Latin scripts~\cite{marchisio2024does}. These adverse affects also arise in QAT-trained models; even though final test accuracy is similar, performance of quantized LLMs can be significantly worse on complex tasks such as multi-turn dialog on standard benchmarks~\cite{dutta2024accuracy}.   

\paragraph{Lossless Compression of LLM Weights}
Compared to the large body of work on 
lossy compression of LLM weights, there is relatively little work on lossless compression. 
An early paper~\cite{han2015deep} proposed Huffman compression of 
convolutional neural network (CNN) weights, naively compressing entire 16-bit weights which incurs large performance overheads. For this reason, they do not actually implement Huffman coding, and instead only use 
run-length encoding (RLE) of sequences of zero weights. RLE is subsequently 
implemented in several other works, especially for CNNs that have sparse weight tensors~\cite{chen2016eyeriss}. 
In contrast, \textsc{Huff-LLM} proposes a lightweight 
hardware-friendly implementation of Huffman coding, 
integrates within 
systolic array and vector LLM accelerator architectures, demonstrating 
substantial performance and energy benefits. 

Two recent works have addressed Huffman compression for LLMs: \cite{hershcovitch2024zipnn} propose to compress the exponent bits of weights via Huffman coding
to reduce the cost of storing and downloading LLMs on cloud servers. For FP16 and BF16 models, they are able to achieve 17 - 33\% compression with higher compression ratios coming from BF16 models due to the larger number of exponent bits. 
\cite{hao2024neuzip} use a similar approach by applying asymmetric numeral systems (ANS), an entropy coding method,  to the exponent bits. In addition, they load the compressed weights to the GPU/TPU and thus achieve memory savings over ~\cite{hershcovitch2024zipnn} during inference. They are able to achieve 33\% lossless compression on BF16 models, but also suffer a 33\% inference slow down due to decompression.


Note that other lossless compression schemes exist. RLE, mentioned previously, 
exploits spatial correlations between inputs by encoding a sequence of identical
weights as the weight value followed by the number of occurrences. 
LZW, a more sophisticated variant, exploits commonly occurring patterns in the input data~\cite{welch1984technique}. Both can be implemented synergistically after Huffman coding of individual weight values. We leave an evaluation of these methods as future work, but note that these incur additional hardware costs.

\subsection{Hardware Accelerators for LLM Inference}
\label{sec:systolic_array}
\begin{figure}[h] 
    \centering
    \includegraphics[width=\linewidth]{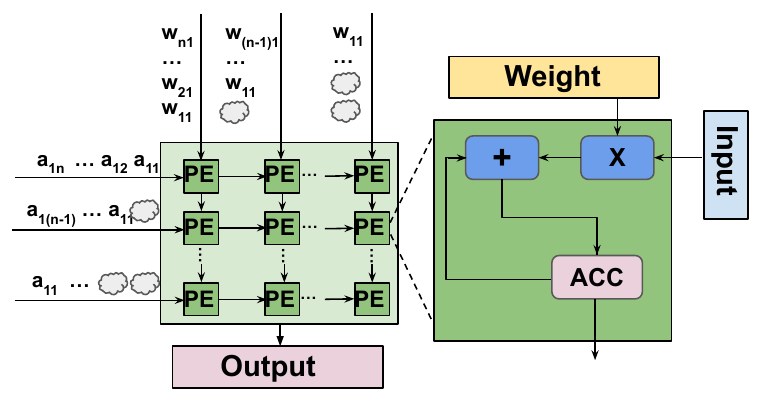} 
    \caption{Systolic array diagram with a detailed look at the PE. Weights and activations are sent to the PEs at every clock cycle. Bubbles indicate delays which are necessary to maintain accuracy of the computations performed by the output stationary architecture. Colors are associated to different operands. Weights are yellow, inputs are light blue, and partial-sum/outputs are pink}
    \label{fig:systolic_array_diagram}

    \vspace{1em} 

    \includegraphics[width=\linewidth]{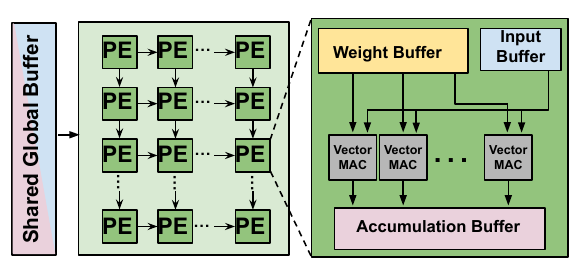} 
    \caption{Simba-like architecture diagram with a detailed look at the PE. Colors are associated to different operands. Weights are yellow, inputs are light blue, and partial-sum/outputs are pink.}
    \label{fig:simba-like_arch_diagram}
\end{figure}

\paragraph{Systolic Array Architectures} 
The systolic array (SA) architecture, shown in Figure~\ref{fig:systolic_array_diagram} consists of an array of processing elements (PEs) that perform multiply-and-accumulate (MAC)
operations, 
surrounded by on-chip buffers for data storage. Weights and activations are fetched from the weight buffer and activation buffer to the PEs, respectively. 
Data is streamed in from these buffers in a 
highly synchronized fashion such that each PE 
computes the dot product of a row of activations with a column of weights. 
However, 
it is crucial to maintain an uninterrupted data flow for correct performance of the systolic array, as
any stalls or bubbles cause either incorrect computation or incur large performance penalties~\cite{sa-pipeline}.
This underscores the necessity of a Huffman decoder that operates without stalling the data stream (See Fig.~\ref{fig:systolic_array_diagram}). 
Note that the description above is for an "output stationary" (OS) systolic array. A slightly different architecture, referred to as weight stationary (WS) stores weights inside each PE and only streams in activations such that the output of each column produces a dot-product of a column of weights with activations. 


\paragraph{Simba Vector Architectures} 

To ensure the generality of our approach, we extended our evaluations to a parallel vector-processing optimized accelerator based on NVIDIA's production-tested NVIDIA Deep Learning Accelerator (NVDLA) architecture~\cite{nvdla, simba}. The hardware model incorporates NVDLA dataflow-optimizations that reduce data-movement for transformers~\cite{nvidia_xformer}. Our evaluations use a single chiplet with an array of 16$\times$16 Processing Elements (PEs) and a shared global buffer for activation storage (See Fig.~\ref{fig:simba-like_arch_diagram}). Each PE features dedicated local scratchpads for weights, inputs, and partial sums, along with vector multiply-accumulate (VMAC) units for parallel computation. The architecture is optimized for a `local-weight-stationary' dataflow (where weights remain fixed in local memory to minimize data movement)~\cite{magnet}, operates at a nominal frequency of 2 GHz, and connects to external LPDDR4 DRAM via a 128 GB/s interface. For ease of reproducibility, detailed hardware specifications are provided in Table~\ref{Simba_specs}.


\paragraph{Lossless Compression in Hardware}
Prior work neural networks accelerators have applied lossless compression 
methods like RLE or sparse coding techniques, but have not implemented full end-to-end entropy coding methods like Huffman coding, in large measure due to its perceived costs. Prior work has implemented lossless compression tailored on CPUs for workloads relevant to general-purpose computing benchmarks.
Bit-Plane Compression (BPC)~\cite{kim2016bit}, for example, introduces a novel compression algorithm to compress homogeneously typed memory blocks. BPC transforms the data and then applies run-length encoding and a frequent pattern encoding to compress the data. 
Buddy Compression~\cite{choukse2020buddy} uses BPC to connect GPU device memory to a ``larger-but-slower buddy memory''. Using a high-bandwidth interconnect between these two memories, they are able to send compressed data to the GPU memory while putting any data that doesn't fit on the GPU into the buddy memory. Selective Memory Compression~\cite{nihaal2024selective} introduces a memory compression scheme that aims to reduce page thrashing by gradually compressing read-only pages. 

\section{The \textsc{Huff-LLM} Scheme}

\subsection{Hardware-Friendly Huffman Compression}
A key challenge with hardware implementations of
Huffman decompression (and other 
entropy coding schemes) is that it is a variable length
code. A naive hardware implementation of Huffman decompression can read a fixed number of 
code-word bits 
in each clock cycle, and output a decompressed source symbol
when a match is found. Thus, when used to decompress a vector of 
weights, this scheme would output valid weights in 
some clock cycles and ``bubbles" (indicating that absence of a valid weight) in others when no match is 
found. As noted in Section~\ref{sec:systolic_array}, neural network 
accelerators like systolic arrays are carefully synchronized 
and require weights (and activations) to be output in each clock cycle for correct operation. 
Dealing with bubbles incurs a large
performance penalty since the entire array needs to be stalled anytime a bubble is encountered, or requires
complex control logic, extra buffering and a potential redesign of the accelerator logic. 

\begin{figure}
    \centering
\includegraphics[width=0.9\columnwidth]{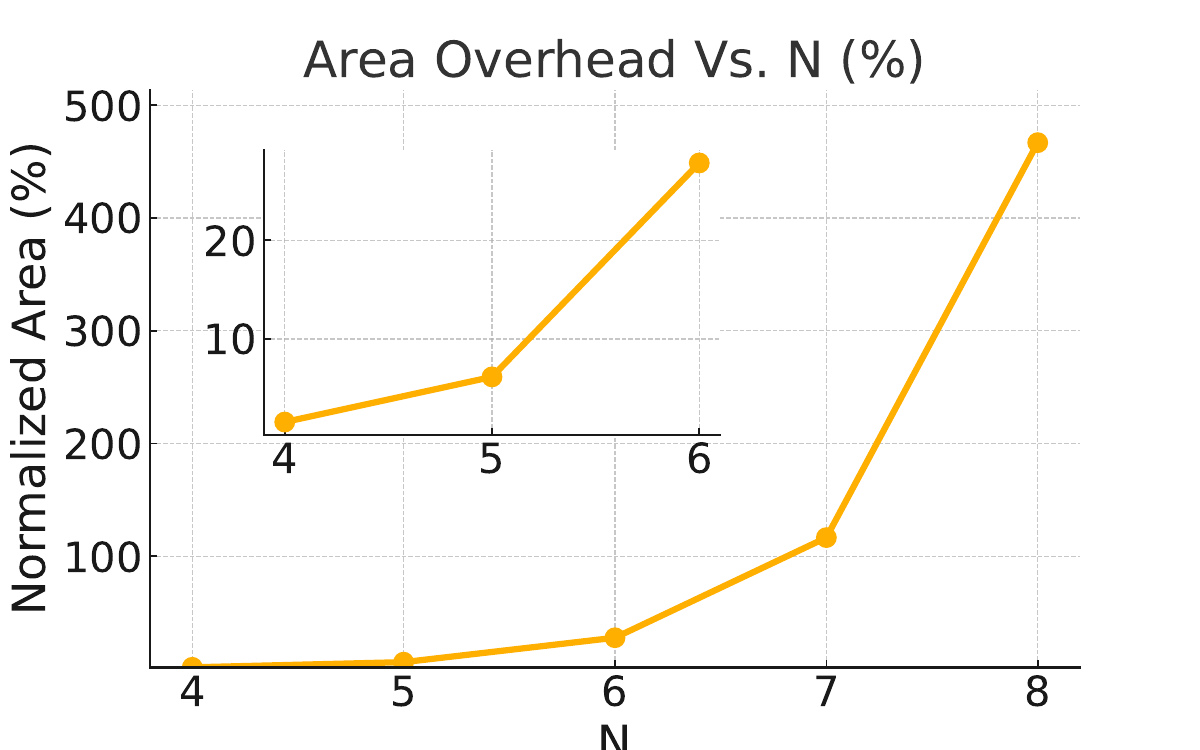} 
    \vspace{-0.15in}
    \caption{Area overhead of a CAM lookup for a single-cycle N-bit Huffman decoder normalized to a column of 128 FP16 multipliers, both clocked at 1 GhZ. Area overheads of Huffman decoding grow quickly, leaving only N=\{4,5\} as viable options.} 
    \label{fig:huffman_cost} 
\end{figure}

On the other hand, a Huffman decoder that outputs a new weight value (or source symbol) in 
each clock cycle enables easy ``plug-and-play" integration into existing neural network accelerators since the decoder can be added as an extra stage in the pipeline. 
Single cycle Huffman decoding, however, introduces a new challenge: an input codeword must be matched against \emph{all} possible $2^{N}$ codewords (assuming $N$-bit source symbols). In hardware, this logic is implemented using 
a content-addressable memory (CAM). However, 
CAMs have high hardware costs, and are typically limited to 32- or 64-entries in applications where single-cycle CAM look-ups are needed. 
Figure~\ref{fig:huffman_cost} plots the CAM overheads 
for 4 to 8 bit source symbols normalized to a column of 
128 FP16 multipliers as reference 
(as we will see shortly, a single Huffman decoder will be shared across a column/row of MAC units). We see that 
overheads are $~6\%$ for N=5 bits, but balloon quickly for 
larger values of N. 

\begin{table}[t]
    \centering
    \label{tab:entropy_splits}
        \begin{tabular}{@{}ccc@{}} \toprule
        Split & Entropy (Bits/Param) & Total Bits/Param\\ \midrule
        16 & 10.54 & 10.54\\
        8-8 & 5.54, 5.03 & 10.57\\
        1-5-5-5 & 1.00, 2.60, 4.97, 2.04 & 10.61\\
        4-4-4-4 & 2.14, 3.91, 4.00, 1.34 & 11.09 \\
        \bottomrule
        \end{tabular}
        \caption{\sl Entropy is calculated for each set of bits as defined by the split. Adding all entropy values together will give the average bits/parameter for the entire weight matrix.}
\end{table}

\begin{figure}
    \centering
\includegraphics[width=0.8\columnwidth]{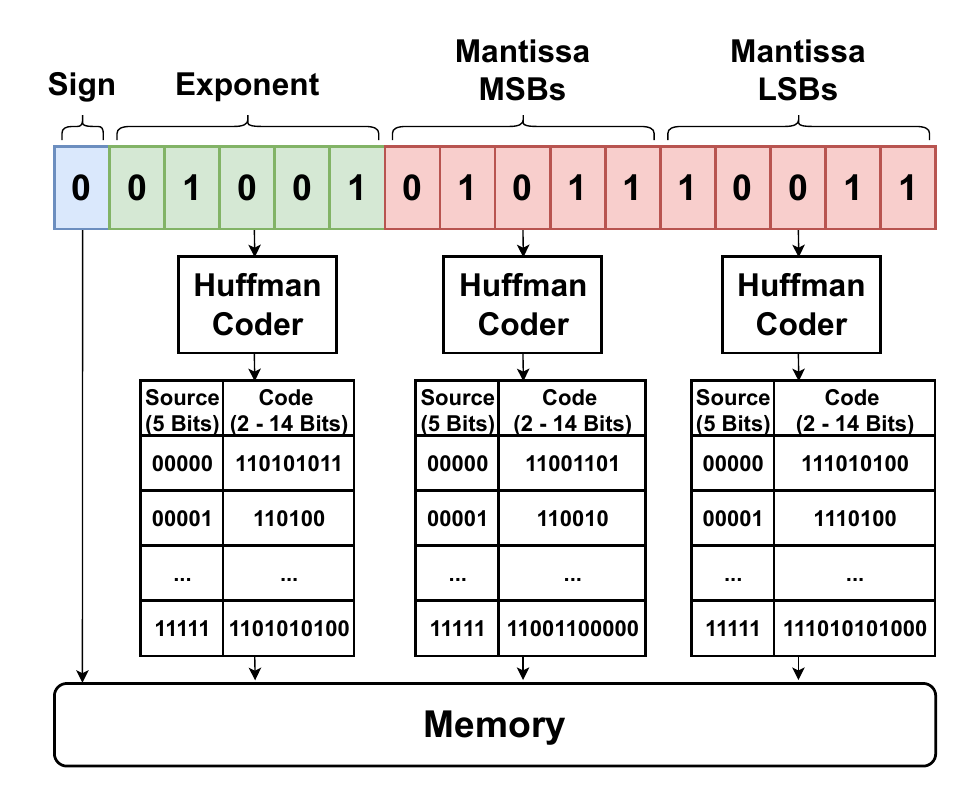} 
    \vspace{-0.1in}
    \caption{Our Huffman Compression method follows these steps for every parameter. It breaks a FP16 number into 4 groups of bits. The sign bit remains uncompressed. The exponent, and mantissa bits are sent through a Huffman Coder to be compressed. They are then stored in memory until they are needed for inference.} 
    \label{fig:huffman_flow} 
\end{figure}

Table~\ref{tab:entropy_splits} shows the entropy of 
Llama-3-8B FP16 weights is 10.54 bits/parameter; 
of course, as we have already observed in Figure~\ref{fig:huffman_cost}, N=16-bit Huffman decompression is infeasible.   
Interestingly, we find that if, instead of Huffman compressing FP16 weights directly, 
we separately compress the 5-bit exponent, the 5 higher-order and 5 lower-order bits of the mantissa, the total entropy 
is only slightly larger at 10.61 bits/parameter. Note that in this scheme, we do not compress the sign bit. We refer to this as $\{1,5,5,5\}$ compression. Also shown in the table is the entropy for $\{8,8\}$ which also has a similar entropy of
10.57 bits/parameter, but is also infeasible from a hardware 
standpoint. 
Based on this analysis, we Huffman compress our weights 
using  $\{1,5,5,5\}$ Huffman compression, as shown in Figure~\ref{fig:huffman_flow}. 
Note that Huffman compression of LLM weights is performed 
only once and can be done offline on a CPU. 
Compressed weights are stored in memory, loaded into on-chip weight buffers
in compressed format, and decompressed only when needed. 
Next, we describe our implementation of 
Huffman decompression in our baseline hardware 
accelerators.

\subsection{Hardware Integration of Huffman Decoders}
\label{para:systolic_decoder_integration}

\begin{figure}
    \centering
\includegraphics[width=0.48\textwidth]{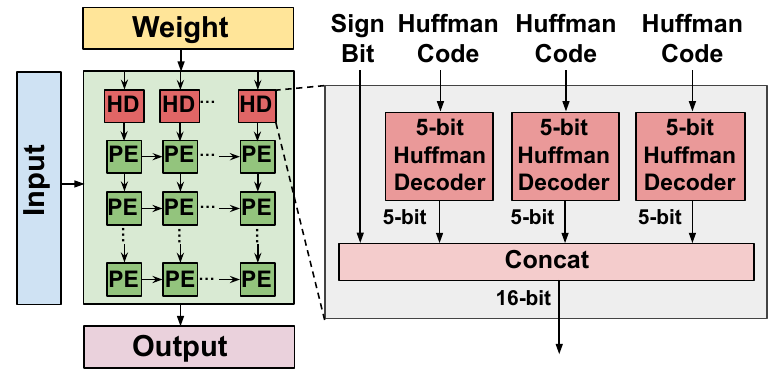}
    \vspace{-0.25in}
    \caption{Systolic array and Huffman Decoder integration.} 
    \label{fig:pe_hd} 
    \vspace*{-0.3in}
\end{figure}

Fig~\ref{fig:pe_hd} shows how Huffman decoders are integrated 
within the baseline systolic array architecture described in 
Section~\ref{sec:systolic_array}. As noted previously, weights are streamed into the systolic array 
from the weight buffer, one weight per clock cycle. In 
\textsc{Huff-LLM}, weights are stored in the weight buffer 
in compressed form.

To enable computation, the Huffman-compressed weight data must be decoded before being fed into the systolic array. As depicted in Fig~\ref{fig:pe_hd}, a row of Huffman decompressors (HD) are inserted between the weight buffer's output and the first row of processing elements (PEs) in the systolic array. Each HD module contains three 
5-bit Huffman decoders. These decoders process the compressed data stream, decoding it into their respective parts of the 
5-5-5 split decompressed data. To enable this, each column's weight buffer is partitioned into three equally sized banks; each bank holds compressed weights from one of the three splits. The sign bit is passed through directly without modification. Finally, the sign bit, exponent, and mantissa bits are concatenated to reconstruct the 16-bit decompressed weight data. 

Figure~\ref{fig:pe_hd_appdx} shows the design of  
each 5-bit HD that enables it to output a decompressed 
weight in a single cycle. 
A 32-bit register holds compressed
values fetched from the (compressed) weight buffer, along with a Start pointer ($S$) that points to the beginning of the 
current codeword. Assuming $L_{max}$ is the number of bits in the longest codeword (note that $L_{max}$ is known in advance, and $L_{max}<32$ for any valid codebook),
bits at positions $S$ to $S+L_{max}-1$ are used 
to match against all codewords stored in a 32-entry CAM. 
Each CAM entry also stores the 5-bit source symbol corresponding to each codeword and the codeword's length, $L$. 
This source symbol is provided as an output from the decompressor
and 
the start pointer is updated to $S \leftarrow S+L$. 
Finally, $L$ bits are read from the weight buffer into the 
codeword register. In practice, a larger codeword register, 
say a 64-bit register can be used with the advantage that 
the weight buffer would only need to be accessed when 
fewer than 32 valid bits are left in the register.

We use the same HD design for the Simba-like vector architecture shown in Figure~\ref{fig:simba-like_arch_diagram}. The HD blocks are inserted between the 
distributed weight memory and vector MAC units. Since, as described, the HD blocks output a new decompressed weight 
per clock cycle without any bubbles or stalls, the throughput/performance of the accelerator is not impacted and 
no changes to the design are needed.

\section{Experimental Setup}

We now describe our experimental setup, including 
architectural parameters of our two baseline neural network accelerators, simulation methodology to estimate performance and energy with and without \textsc{Huff-LLM}, 
LLMs and the datasets on which they are evaluated.

\paragraph{Systolic array settings and simulation.}
Table~\ref{tab:sim_spec} shows the architectural parameters 
of our baseline systolic array architecture, reflective
of an edge tensor processing unit (TPU) similar to
the Google Coral edge device~\cite{suryavansh2020google}. 
The simulated architecture has a peak performance of 16~TOPs at 16b float precision, and was evaluated with 64 GBps and 128 GBps memory bandwidth.
We model both output stationary (OS) and weight stationary (WS) architectures, as described in Section~\ref{sec:systolic_array}. 
We model the performance and energy of this architecture using a methodology similar to STAR-Sim~\cite{starsim} and SCALE-Sim~\cite{scalesim} that are both widely used to model systolic array architectures.
(more details in Appendix~\ref{appendix:star-sim}). However, STAR-Sim and SCALE-Sim focus on modelling convolution operations while a majority of computations in LLMs are matrix multiplications~\cite{Wang2020SpAttenES}. 
We modify SCALE-Sim for faster matrix multiplication simulations. 


\paragraph{Simba architecture settings and simulation.}  
We tabulate the architectural specifications of the evaluated Simba-like architecture in Table~\ref{Simba_specs}. Parameters were selected based on designs available in Nvidia Research's Timeloop/Accelergy repository\footnote{https://github.com/Accelergy-Project/timeloop-accelergy-exercises}. 
The simulated architecture has a peak performance of 4~TOPs at 16b float precision, in-line with mobile NPUs~\cite{samsung_npu_isca}, and was evaluated with 64 GBps and 128 GBps memory bandwidth, similar to the systolic array. 
Timeloop is an accelerator performance estimation tool developed by Nvidia~\cite{timeloop}. Timeloop can model various scheduling strategies, and estimate how they impact the energy and latency of a computation. We use Timeloop’s hybrid search across all our experiments to minimize inference latency first and energy
for mappings with the same latency.
Energy costs are estimated via Accelergy via access counts generated from Timeloop. 
This approach is used to calculate energy for larger memories through Cacti \cite{balasubramonian2017cacti} and smaller components such as address generators and register files using figures included in Aladdin \cite{aladdin}. 

\begin{table}[t]
    \centering
    
    \resizebox{0.9\linewidth}{!}{%
        \begin{tabular}{@{}ccc@{}} \toprule

        Tech node & \multicolumn{2}{c}{$16nm$} \\ 
        Systolic Array Size & \multicolumn{2}{c}{$128 \times 128$ PE}    \\ 
        PE Frequency & \multicolumn{2}{c}{$1GHz$}    \\ \midrule
        DRAM Bandwidth & $64GB/s$ & $128GB/s$    \\ 
        Weight Buffer Size & \multicolumn{2}{c}{$16KB$}    \\ 
        Activation Buffer Size & \multicolumn{2}{c}{$8KB$}    \\ 
        Accumulator Buffer Size & \multicolumn{2}{c}{$4KB$}    \\ 

        Dataflow & WS & OS    \\

        \bottomrule
        \end{tabular}
    }

        \caption{\sl Systolic Array Architecture Specifications.}
        \label{tab:sim_spec}
\end{table}

\begin{table}[t]
    \centering
    \label{tab:simba_arch_spec}
    \resizebox{\linewidth}{!}{%
        \begin{tabular}{@{}lccc@{}} \toprule

        \multirow{3}{*}{Chip} & Tech node & \multicolumn{2}{c}{$16nm$} \\
                                  & PE Frequency & \multicolumn{2}{c}{$2GHz$} \\
                                  & Number of PEs & \multicolumn{2}{c}{$16$} \\
        \midrule
        \multirow{5}{*}{PE} & DRAM Bandwidth & $64GB/s$ & $128GB/s$ \\
                              & Weight Buffer Size & \multicolumn{2}{c}{$32KiB$} \\
                              & Input Buffer Size & \multicolumn{2}{c}{$8KiB$} \\
                              & Accumulator Buffer Size & \multicolumn{2}{c}{$3KiB$} \\
                              & Number of Vector MACs & \multicolumn{2}{c}{$8$} \\
                              & Vector MAC Width & \multicolumn{2}{c}{$8$} \\
        \bottomrule
        \end{tabular}
    }
    \caption{\sl Simba Architecture Specifications.}
    \label{Simba_specs}
\end{table}





\paragraph{Benchmarks and Evaluated LLMs.}
Benchmarks such as MMLU~\cite{hendrycks2020measuring} are typically used to measure LLM capabilities. Works that focus on lossy compression (such as quantization) often use benchmark performance to show how much information was lost in the compression process. However, since \textsc{Huff-LLM} is a lossless compression method, the compressed LLM maintains exactly the same accuracy as the original model by construction. 

Alternatively, we can view LLM benchmarks from the perspective of their input context size. Different benchmarks have different average lengths of their inputs. For example, Arc-Easy~\cite{clark2018think} has an average length of approximxately 42 input tokens, whereas MMLU has an average length of approximately 92 input tokens. Therefore, we use benchmarks to test how \textsc{Huff-LLM}'s optimizations are impacted by various input token lengths. 

Our performance estimation system uses the average token length of a benchmark query, rather than the actual benchmark questions. We report the average token length (as determined by the Llama-3-8B tokenizer) of each benchmark used in Table~\ref{tab:benchmark}.

\begin{table}[!t]
    \centering
    \label{tab:benchmark}
        \begin{tabular}{@{}cccc@{}} \toprule
        Benchmark & ArcEasy & MMLU & Winogrande \\ \midrule
        Avg Input Tokens & 42 & 92 & 25\\
        Standard Deviation & 20 & 92 & 4 \\
        \bottomrule
        \end{tabular}
        \caption{\sl Average input token length for benchmarks. Tokens are generated with Llama-3-8B's tokenizer.}
\end{table}

We perform compression tests and hardware simulations on various notable LLM families. We included the Llama~\cite{dubey2024llama}, OPT~\cite{zhang2022opt}, Qwen~\cite{yang2024qwen2}, and Vicuna~\cite{vicuna2023} model families to show how \textsc{Huff-LLM} performs on different model architectures. In addition, we examine model sizes ranging from 3B to 13B parameters to see how model size impacts our compression scheme.
\vspace{-0.2in}
\section{Experimental Results}

\subsection{Compression Experiments}
We apply our Huffman compression method to FP16 variants of popular LLM families such as Llama, OPT, Qwen, and Vicuna. The total compression ratio is calculated by averaging the compression ratio of the attention and mlp weight matrices. These results can be found in Table~\ref{tab:comp_ratios}. We notice that the Llama and Vicuna model families have similar compression ratios even at different model sizes. However, we also see that OPT and Vicuna have a notably smaller compression ratio. This suggests that there may be factors during the training stage that lead to certain distributions (and thus lossless compressibility) of the weights. 

BF16 models are also very popular for inference. Therefore, we adapt our Huffman Compression method to work with BF16 models as well. We apply the same idea when splitting the bits. We find that the bits can be split 1-4-4-7. The seven mantissa bits at the end show little to no compressibility (in contrast to FP16). The sign bit remains uncompressed, and the exponent bits are split into two groups of four. We compress various models using our method and report the ratios in Table~\ref{tab:comp_ratios}. We notice that all model families have a similar compression ratio in BF16. This is in contrast to FP16 where Vicuna and OPT had notably lower compression ratios. This could arise from the conversion process to BF16. We see that the compressibility of the mantissa (in FP16) has moved entirely to the exponent (in BF16). Therefore, in models where the FP16 compressibility is low (Vicuna), we are likely seeing higher compressibility in BF16 due to the larger number of exponent bits. 

\begin{table}[t] 
    \centering
    \label{tab:comp_ratios}
    \resizebox{\columnwidth}{!}{%
        \begin{tabular}{@{}cccccc@{}} \toprule
        \textbf{Model Name} & \multicolumn{2}{c}{\textbf{FP16}} & \multicolumn{2}{c}{\textbf{BF16}} \\ \midrule
        &Bits/Param  & Ratio & Bits/Param & Ratio\\ 
        \cmidrule(lr){2-3} \cmidrule(lr){4-5}
        Llama-3.2-3B & 10.96 & 1.46 & 11.68 & 1.37\\
        Llama-3-8B & 10.96 & 1.46 & 11.68& 1.37 \\ 
       Llama-2-13B & \textcolor{blue}{\textbf{10.88}} & \textcolor{blue}{\textbf{1.47}} & 11.59 & 1.38\\ \hline
          OPT-2.7B & 13.68 & 1.17 & 11.68&1.37\\
        OPT-6.7B & \textcolor{red}{\textbf{13.78}} & \textcolor{red}{\textbf{1.16}}& 11.68 &1.37 \\
         OPT-13B & 13.68 & 1.17 & 11.59&1.38\\ \hline
         Qwen-2.5-3B & 10.96&1.46 & 11.68 & 1.37\\
        Qwen-2.5-7B & 10.96 &1.46 &11.68&1.37\\ \hline
        Vicuna-7B & 13.68 & 1.17& 11.59&1.38\\
        Vicuna-13B &  13.68&1.17 & 11.59&1.38\\ 
        \bottomrule
        \end{tabular}
    }

        \caption{\sl Compression ratio is calculated as an average of all weight matrices in each model. Bits/Param is calculated by dividing the uncompressed Bits/Param (16) by the compression ratio. Highest and lowest ratios are highlighted.}
        \vspace{-0.2in}
\end{table}

\subsection{Hardware Results}

\paragraph{Latency and Energy Savings}
Tables~\ref{tab:main_result_latency} and~\ref{tab:main_result_energy} present the latency and energy savings achieved when applying \textsc{Huff-LLM} to various FP16 models. We explore the Llama and OPT model families because they represent the best and worst case scenario for \textsc{Huff-LLM}. Since compression ratio plays the largest role in determining these simulation results, we omit Qwen and Vicuna results due to their similarity with Llama and OPT respectively. In addition, full OPT results can be found in the Appendix in Section~\ref{sec:opt_results}.

The \textsc{Huff-LLM} compression scheme leads to significant latency improvements, ranging from 26\% to 31\% for LLaMA models and 13\% to 15\% for OPT models. In terms of energy savings, LLaMA models also achieve notable gains of 16\% to 26\%, whereas OPT models see improvements of 3\% to 10\%. Additionally, reducing the DRAM bandwidth from 128 GB/s to 64 GB/s slightly enhances latency improvements by 2\% to 3\% but reduces energy savings by a similar margin. A comparable trend is observed for the OPT models. For Systolic Array architectures we see higher energy savings than for Simba architectures. This is due to the smaller weight buffer and increased reliance on DRAM access; the Systolic Array benefits more from \textsc{Huff-LLM} because a larger portion of its overall energy consumption comes from memory-related operations.

\begin{table*}[h!]
\centering
\begin{adjustbox}{max width=\textwidth}
\begin{tabular}{cccccccccc}
\hline

\textbf{Benchmark} & \textbf{Bandwidth}          & \multicolumn{2}{c}{\textbf{Llama 2-13B}} & \multicolumn{2}{c}{\textbf{Llama 3-8B} }& \multicolumn{2}{c}{\textbf{Llama 3.2-3B}} & \multicolumn{2}{c}{\textbf{OPT-13B}}\\ \hline
& & Systolic Array & Simba & Systolic Array & Simba & Systolic Array & Simba & Systolic Array & Simba\\
\cmidrule(lr){3-4} \cmidrule(lr){5-6} \cmidrule(lr){7-8} \cmidrule(lr){9-10}
\multirow{2}{*}{MMLU} & 64 GB/s             & 29.41\%            & 28.18\% & 29.40\%           &  31.33\%  & 28.50\%  & 31.07\% &  14.77\% & 14.05\%           \\ 
                      & 128GB/s             & 27.17\%    &    \textcolor{red}{\textbf{26.97\%}}  & 28.97\%  & \textcolor{red}{\textbf{27.16\%}}         &  28.22\%   & \textcolor{red}{\textbf{26.23\%}} &14.01\% & \textcolor{red}{\textbf{13.33\%}}              \\ \hline
\multirow{2}{*}{Winogrande} & 64 GB/s        & 29.86\%              &31.11\% &29.86\%           &  31.11\% & 29.44\% & \textcolor{blue}{\textbf{31.38\%}}  & \textcolor{blue}{\textbf{15.16\%}} & 14.29\%            \\ 
                      & 128 GB/s             & 27.63\%            & 29.60\% & 27.63\%           &  29.58\% & 27.20\%   & 29.58\%  &14.41\% & 13.33\%          \\ \hline
\multirow{2}{*}{ArcEasy} & 64 GB/s          & 29.74\%            & \textcolor{blue}{\textbf{31.40\%}} & 29.74\%            & \textcolor{blue}{\textbf{31.35\%}} & 29.20\% & 31.33\%  & 15.06\% & 14.29\%            \\ 
                      & 128 GB/s             & 27.51\%            & 29.60\% & 27.51\%           &  29.58\% & 26.95\% & 29.53\% & 14.30\% & \textcolor{red}{\textbf{13.33\%}}             \\ \hline
\end{tabular}
\end{adjustbox}
\caption{Latency savings achieved when applying Huff-LLM to different FP16 models. Savings are simulated for Systolic Arrays  and Simba with weight stationary (WS) architectures. Highest and lowest improvements are highighted.}
\label{tab:main_result_latency}
\end{table*}

\begin{table*}[h!]
\centering
\begin{adjustbox}{max width=\textwidth}
\begin{tabular}{cccccccccc}
\hline

\textbf{Benchmark} & \textbf{Bandwidth}          & \multicolumn{2}{c}{\textbf{Llama 2-13B}} & \multicolumn{2}{c}{\textbf{Llama 3-8B} }& \multicolumn{2}{c}{\textbf{Llama 3.2-3B}} & \multicolumn{2}{c}{\textbf{OPT-13B}}\\ \hline
& & Systolic Array & Simba & Systolic Array & Simba & Systolic Array & Simba & Systolic Array & Simba\\
\cmidrule(lr){3-4} \cmidrule(lr){5-6} \cmidrule(lr){7-8} \cmidrule(lr){9-10}
\multirow{2}{*}{MMLU} & 64 GB/s             &     23.76\%        & \textcolor{red}{\textbf{16.12\%}} &    23.75\%        &  \textcolor{red}{\textbf{16.47\%}}  &        22.78\%  &\textcolor{red}{\textbf{16.67\%}}  & 9.14\% & 3.40\%  \\ 
                      & 128GB/s             &    25.55\%       & 19.79\%  &    25.54\%      &     19.33\%&  24.59\%   & 19.72\%   & 9.71\%  & 6.41\%     \\ \hline
\multirow{2}{*}{Winogrande} & 64 GB/s        &   24.25\%          & 18.98\%&   24.25\%       & 18.68\%  &      23.80\% &18.82\% & 9.56\% & 4.82\%     \\ 
                      & 128 GB/s             &    \textcolor{blue}{\textbf{26.03\%}}        &  20.30\%&    \textcolor{blue}{\textbf{26.02\%}}      &   19.56\%&   \textcolor{blue}{\textbf{25.58\%}}   & 19.98\%  & \textcolor{blue}{\textbf{10.13\%}}  & 6.05\%    \\ \hline
\multirow{2}{*}{ArcEasy} & 64 GB/s          &    24.12\%         & 17.44\% &     24.12\%      & 18.22\% &     23.53\% &17.74\%   & 9.45\%  & \textcolor{red}{\textbf{3.31\%}}   \\ 
                      & 128 GB/s             &    25.90\%      & 20.78\% &    25.90\%    &  20.21\% &   25.32\% &19.61\%  & 10.02\%  & 5.00\% \\ \hline
\end{tabular}
\end{adjustbox}
\caption{Energy savings achieved when applying Huff-LLM to various FP16 models. Savings are simulated for Systolic Arrays and Simba with weight stationary (WS) architectures. Highest and lowest improvements are highighted.}
\label{tab:main_result_energy}
\end{table*}

\begin{figure}[!t]
    \centering
    \includegraphics[width=0.8\linewidth]{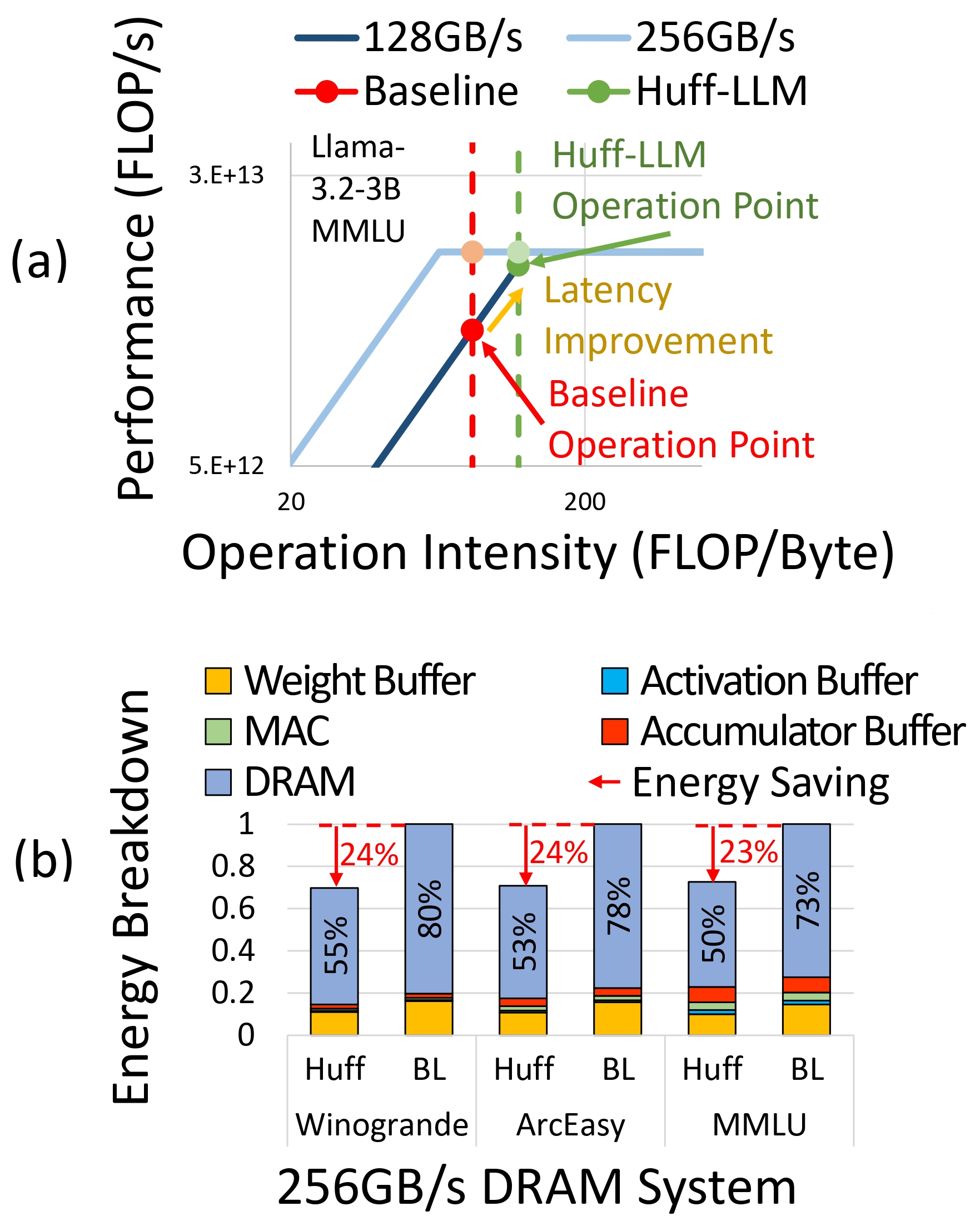}
    \vspace{-0.1in}
    \caption{(a) Roofline plot of Systolic Arrays with 128GB/s and 256GB/s DRAM bandwidth. Dashed lines show the baseline and Huff-LLM models, with intersections marking operational points. (b) Energy breakdown of Huff-LLM and baseline model on the Systolic Array with 256GB/s DRAM bandwidth.} 
    \label{fig:starsim-roofline}
\end{figure}

The results thus far are for weight stationary (WS) architectures. 
Table~\ref{tab:output_stationary} shows latency reductions and energy savings
for output stationary (OS) systolic arrays;
we observe Huff-LLM consistently achieves larger improvements in output stationary architectures because the output stationary system has less weight reuse. This leads to higher weight movement overhead which can be reduced by Huff-LLM.

\paragraph{Area Overheads} To estimate the area overheads of the proposed Huff-LLM scheme, we implemented 
a 5-bit Huffman Decoder and a systolic array PE in Verilog (a hardware description language) and synthesized these blocks for a Global Foundries 12nm (GF12) technology. 
Each PE has an area of 484$um^2$ while 
an HD with an empirically determined
$L_{max}=12$ is 1199.3$um^2$. For a 128$\times$128 systolic array, the area overhead of Huff-LLMs is $6.13\%$. The 
area overhead at the full chip level would be even lower since we have not accounted for on-chip buffers in the denominator. Custom, highly-optimized CAM structures~\cite{yue202415} can lower overheads further.

\section{Discussion}


Higher DRAM bandwidth diminishes the latency savings achieved by Huff-LLM, as weight movement accounts for a smaller fraction of the overall processing time. Additionally, as DRAM bandwidth increases, the system becomes more compute-bound. To evaluate the limits of Huff-LLM, we configure the Systolic Array with a DRAM bandwidth of 256GB/s and simulate its performance.
To shed further light on Huff-LLM's performance gains,
Figure~\ref{fig:starsim-roofline}(a) presents a roofline plot~\cite{roofline2008david} for Systolic Arrays executing the MMLU task on the Llama-3.2-3B model. Weight compression 
results in larger FLOPs/Byte since we need to fetch Bytes from memory, resulting in
higher performance (FLOP/s) for 128 GB/s bandwidth. 
However for an even higher 
256 GB/s bandwidth, even the baseline model 
runs at peak performance, so compression will not reduce latency further (this is true for even 8-bit quantization). 
Despite the reduced latency improvement in higher bandwidth systems, Huff-LLM \emph{still lowers energy consumption} by reducing memory access operations. Figure~\ref{fig:starsim-roofline}(b) illustrates the energy breakdown for Huff-LLM and the 16-bit baseline on the 256GB/s system, using the same MMLU task on the Llama-3.2-3B model. While system speed remains unchanged, Huff-LLM achieves a $24\%$ reduction in energy consumption.
This is largely because of the dominant energy costs of fetching data from memory, compared to other on-chip costs. 

\section{Conclusion}
In this work, we propose Huff-LLM, an end-to-end model compression method for LLMs. We observe that Huffman Compression can be applied to subsets of weight parameters with minimal impact on the compression ratio. We use this observation to develop a compression scheme and hardware design that has minimal area overhead and is fast. We show up to 32\% reduction in model size, up to 31\% improvement in inference latency, and up to 26\% reduction in energy cost.

\section*{Impact Statement}

This paper presents work whose goal is to advance the field of  Machine Learning. There are many potential societal consequences of our work, none which we feel must be specifically highlighted here.

\bibliography{example_paper}
\bibliographystyle{icml2025}

\newpage
\appendix
\onecolumn
\section{Additional results}

You can have as much text here as you want. The main body must be at most $8$ pages long.
For the final version, one more page can be added.
If you want, you can use an appendix like this one.  

The $\mathtt{\backslash onecolumn}$ command above can be kept in place if you prefer a one-column appendix, or can be removed if you prefer a two-column appendix.  Apart from this possible change, the style (font size, spacing, margins, page numbering, etc.) should be kept the same as the main body.

\subsection{Analytical Simulator}
\label{appendix:star-sim}
To map the multiplication of input matrix $I$ and weight matrix $W$, with shape of $I_H \times I_W$ and $W_H \times W_W$, we adopted same dataflow-dependent mapping schemes as SCALE-Sim~\cite{scalesim}, where the dimensions of two operand matrices are defined as $S_R \times T$ and $T \times S_C$, as shown in Figure~\ref{fig:pe_hd}. $S_R$ and $S_C$ are the specific workload dimensions mapped to the rows and columns of a systolic array, respectively, and T is the temporal dimension, along which the data are being streamed into the systolic array. Table~\ref{tab:sa_map} summarizes the definition of $S_R$, $S_C$, and $T$ depending on the dataflow configuration. 

\begin{table}[ht]
    \centering
    \label{tab:sa_map}
        \begin{tabular}{@{}cccc@{}} \toprule

        Dataflow & $S_R$ & $S_C$  & $T$   \\ \midrule
        Weight Stationary & $W_H$ &$W_W$ & $I_H$  \\


        Output Stationary & $I_H$ & $W_W$ & $W_H$  \\

        \bottomrule
        \end{tabular}

        \caption{\sl Spatial and temporal mapping of the input matrix (I) with shape of $I_H \times I_W$ and weight matrix (W) with shape of $W_H \times W_W$ to the rows and columns of a systolic array. $I_W$ is equal to $W_H$.}
\end{table}

Since one systolic array may not be sufficient to accommodate the entire matrix computation in common LLM layers, the workload is typically partitioned into "folds"~\cite{scalesim} with respect to the rows ($R$) and columns ($C$) of the PE array. The number of folds along the row dimension ($F_R$) and column dimension ($F_C$) can be calculated as:
\begin{equation}
    F_R=\lceil \frac{S_R}{R} \rceil, F_C=\lceil \frac{S_C}{C} \rceil
\end{equation}

We use the same principle as SCALE-Sim~\cite{scalesim} to model the number of compute cycles as below:
\begin{equation}
    L_{COMP} = (2R+C+T-2)\cdot F_R \cdot F_C
\end{equation}

For modeling of buffer access, we consider the stationary data and streaming data separately. The stationary operand in the systolic array will get updated only after being fully reused by the streaming operand, so the number of read accesses of stationary data is equal to the number of stationary data. In contrast, the streaming data may need to be reloaded by multiple times, for which the reloading count is equal to the number of folds. As such, the number of weight buffer read accesses ($WB_{RD}$) and the number of input buffer read accesses ($IB_{RD}$) can be modeled as below:
\begin{equation}
\label{equation:wb_rd}
    WB_{RD} =
\begin{cases} 
W_H \cdot W_W, & \text{if } WS \\
W_W \cdot W_H \cdot \lceil \frac{I_H}{R} \rceil, & \text{if } OS
\end{cases}
\end{equation}

\begin{equation}
\label{ib_rd}
   IB_{RD} =
\begin{cases} 
W_H \cdot I_H \cdot \lceil \frac{W_W}{C} \rceil, & \text{if } WS \\
I_H \cdot W_H \cdot \lceil \frac{W_W}{C} \rceil, & \text{if } OS
\end{cases} 
\end{equation}

The MAC latency is modeled as the product of the total compute cycles and the cycle time. The memory read/write latency is modeled as the division of read/write data and buffer bandwidth. 

The MAC energy is modeled as the product of the total number of MAC operations and the energy per MAC operation. The memory read/write access energy is modeled as the product of the unit energy per read/write access and the number of read/write accesses. 

\subsection{Additional Results}\label{sec:opt_results}
Full OPT results and output stationary results are included in this section.

\begin{table}[h!]
\centering
\begin{adjustbox}{max width=\textwidth/2}
\begin{tabular}{|c|c|c|c|c|}
\hline

\textbf{Benchmark} & \textbf{Bandwidth}          & \textbf{OPT-13B} & \textbf{OPT-6.7B} & \textbf{OPT-2.7B} \\ \hline
\multirow{2}{*}{MMLU} & 64 GB/s             & 14.77\%             & 14.63\%              & 14.23\%               \\ \cline{2-5}
                      & 128 GB/s             & 14.01\%             & 13.87\%              & 13.47\%               \\ \hline
\multirow{2}{*}{Winogrande} & 64 GB/s        & 15.16\%              & 15.10\%              & 14.93\%               \\ \cline{2-5}
                      & 128 GB/s             & 14.41\%              & 14.35\%              & 14.17\%               \\ \hline
\multirow{2}{*}{ArcEasy} & 64 GB/s          & 15.06\%              & 14.98\%              & 14.75\%               \\ \cline{2-5}
                      & 128 GB/s             & 14.30\%              & 14.22\%              & 13.99\%               \\ \hline
\end{tabular}
\end{adjustbox}
\caption{Latency saving when compressing the OPT model weights from 16-bit to 14-bit. The results are from the Star system simulation. }
\label{tab:starsim-opt-latency}
\end{table}

\begin{table}[h!]
\centering
\begin{adjustbox}{max width=\textwidth/2}
\begin{tabular}{|c|c|c|c|c|}
\hline

\textbf{Benchmark} & \textbf{Bandwidth}          & \textbf{OPT-13B} & \textbf{OPT-6.7B} & \textbf{OPT-2.7B} \\ \hline
\multirow{2}{*}{MMLU} &64 GB/s             & 9.14\%             & 9.00\%              & 8.57\%               \\ \cline{2-5}
                      & 128 GB/s             & 9.71\%             & 9.57\%              & 9.14\%               \\ \hline
\multirow{2}{*}{Winogrande} & 64 GB/s        & 9.56\%              & 9.50\%              & 9.31\%               \\ \cline{2-5}
                      & 128 GB/s             & 10.13\%              & 10.06\%              & 9.88\%               \\ \hline
\multirow{2}{*}{ArcEasy} & 64 GB/s          & 9.45\%              & 9.37\%              & 9.12\%               \\ \cline{2-5}
                      & 128 GB/s             & 10.02\%              & 9.94\%              & 9.69\%               \\ \hline
\end{tabular}
\end{adjustbox}
\caption{Energy saving when compressing the OPT model weights from 16-bit to 14-bit. The results are from the Star system simulation.}
\label{tab:starsim-opt-energy}
\end{table}

\begin{table}[h!]
\centering
\begin{adjustbox}{max width=\textwidth/2}
\begin{tabular}{|c|c|c|c|c|}
\hline
\multirow{1}{*}{\textbf{BenchMark}}   & \multirow{1}{*}{\textbf{Bandwidth}} & \textbf{OPT-13B} & \textbf{OPT-6.7B} & \textbf{OPT-2.7B}      \\ \hline
\multirow{2}{*}{MMLU}        & 64 GB/s                   & 14.05\%      & 14.29\%    & 14.05\%     \\ \cline{2-5} 
                             & 128 GB/s                  & 13.33\%      & 13.35\%    & 13.33\%     \\ \hline
\multirow{2}{*}{Arceasy}     & 64 GB/s                   &14.29 \%      & 14.2\%    & 14.4\%     \\ \cline{2-5} 
                             & 128 GB/s                  & 13.33\%      & 13.29\%    & 13.36\%     \\ \hline
\multirow{2}{*}{Winogrande}  & 64 GB/s                   & 14.29\%      & 14.35\%    & 14.31\%     \\ \cline{2-5} 
                             & 128 GB/s                  & 13.33\%      & 13.35\%    & 13.33\%     \\ \hline
\end{tabular}
\end{adjustbox}
\caption{Latency saving for different benchmarks at 64 GB/s and 128 GB/s bandwidth simulated on timeloop on Simba for OPT models with compressed weights from 16-bit to 14-bit.}
\label{OPT_latency_timeloop}
\end{table}

\begin{table}[h!]
\centering
\begin{adjustbox}{max width=\textwidth/2}
\begin{tabular}{|c|c|c|c|c|}
\hline
\multirow{1}{*}{\textbf{BenchMark}}   & \multirow{1}{*}{\textbf{Bandwidth}} & \textbf{OPT-13B} & \textbf{OPT-6.7B} & \textbf{OPT-2.7B}      \\ \hline
\multirow{2}{*}{MMLU}        & 64 GB/s                   & 3.4\%      & 3.5\%    & 3.38\%     \\ \cline{2-5} 
                             & 128 GB/s                  & 6.41\%      & 6.93\%    & 6.89\%     \\ \hline
\multirow{2}{*}{Arceasy}     & 64 GB/s                   & 3.31\%      & 3.48\%    & 4.02\%     \\ \cline{2-5} 
                             & 128 GB/s                  & 5\%      & 5.78\%    & 6.1\%     \\ \hline
\multirow{2}{*}{Winogrande}  & 64 GB/s                   & 4.82\%      & 5.13\%    & 5.2\%     \\ \cline{2-5} 
                             & 128 GB/s                  & 6.05\%      & 6.23\%    & 6.37\%     \\ \hline
\end{tabular}
\end{adjustbox}
\caption{Energy saving for different benchmarks at 64 GB/s and 128 GB/s bandwidth simulated on timeloop on Simba for OPT models with compressed weights from 16-bit to 14-bit.}
\label{OPT_energy_timeloop}
\end{table}

\begin{table*}[h!]
\centering
\begin{adjustbox}{max width=\textwidth}
\begin{tabular}{cccccccccc}
\hline

\textbf{Benchmark} & \textbf{Bandwidth}          & \multicolumn{2}{c}{\textbf{Llama 2-13B}} & \multicolumn{2}{c}{\textbf{Llama 3-8B} }& \multicolumn{2}{c}{\textbf{Llama 3.2-3B}} \\ \hline
& & Latency & Energy & Latency & Energy & Latency & Energy \\
\cmidrule(lr){3-4} \cmidrule(lr){5-6} \cmidrule(lr){7-8} 
\multirow{2}{*}{MMLU} & 64 GB/s  & 31.00\% &24.37\% & 30.99\% & 24.36\%& 30.11\% & 23.39\%\\ 
                      & 128GB/s & 28.31\% & 26.69\%& 28.30\% & 26.68\%& 27.38\% & 25.75\%          \\ \hline
\multirow{2}{*}{Winogrande} & 64 GB/s  & 31.44\% & 24.85\%& 31.44\% & 24.85\%& 31.03\% & 24.40\%\\ 
                      & 128 GB/s   & 28.76\% & 27.16\%& 28.76\% & 27.16\%& 28.34\% & 26.73\%         \\ \hline
\multirow{2}{*}{ArcEasy} & 64 GB/s & 31.33\% &24.73\% & 31.32\% & 24.72\%& 30.79\% & 24.14\% \\ 
                      & 128 GB/s  & 28.65\% & 27.04\%& 28.64\% & 27.04\%& 28.09\% & 26.47\% \\ \hline
\end{tabular}
\end{adjustbox}
\caption{Latency and Energy savings achieved when applying Huff-LLM FP16 models. Simulations are performed on an output stationary (OS) systolic array architecture.}
\label{tab:output_stationary}
\end{table*}

\subsection{Huffman Decoder Figure}
\begin{figure}[h]
    \centering
    \includegraphics[width=0.7\linewidth]{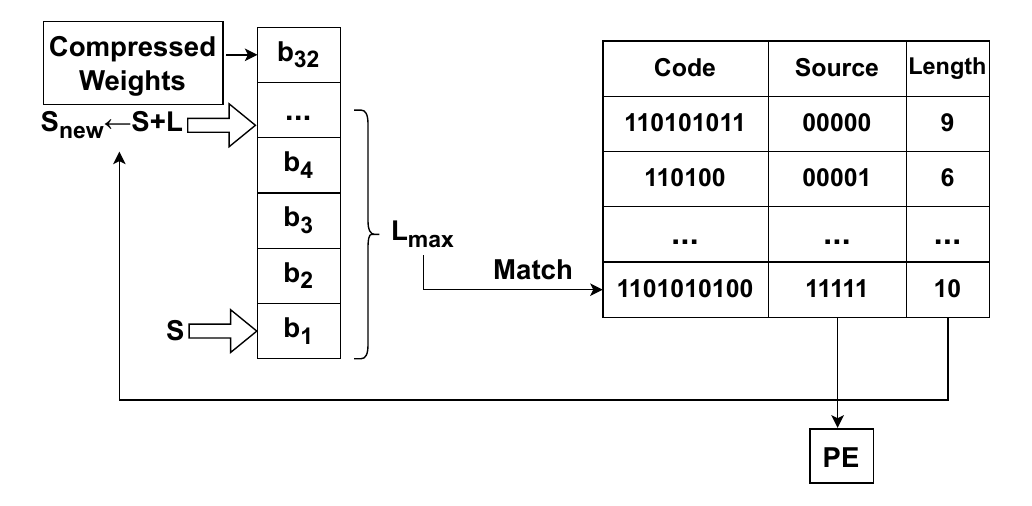}
    \caption{Each Huffman Decoder module follows the process shown in this figure. L$_{max}$ bits are taken from the register and a match is found in the Huffman Table. Afterwards, the decoded source symbol is sent to the PE while the length is sent to update the start position S.}
    \label{fig:pe_hd_appdx}
\end{figure}

\end{document}